\documentclass[runningheads]{llncs}
\usepackage[T1]{fontenc}
\usepackage{graphicx}
\usepackage[hidelinks]{hyperref}
\usepackage{amsfonts}
\usepackage{amsmath}
\usepackage{algpseudocode}
\usepackage{caption}
\usepackage{subcaption}
\usepackage{orcidlink}
\usepackage[acronym,nomain,nonumberlist]{glossaries}
\makeglossaries

\newacronym{drl}{DRL}{Disentangled Representation Learning}
\newacronym{myo}{MYO}{Myocardium}
\newacronym{lv}{LV}{left ventricle blood cavity}
\newacronym{rv}{RV}{right ventricle blood cavity}
\newacronym{ct}{CT}{Computed Tomography}
\newacronym{mri}{MRI}{Magnetic Resonance Imaging}
\newacronym{vmf}{vMF}{von Mises-Fisher}
\newacronym{csd}{CSD}{Content-Style Disentanglement}

\newacronym{dsc}{DSC}{Dice Similarity Coefficient}
\newacronym{lpips}{LPIPS}{Learned Perceptual Image Patch Similarity}
\newacronym{assd}{ASSD}{Average Symmetric Surface Distance}

\usepackage{color}

\begin{document}
\title{Enhancing Cross-Modal Medical Image Segmentation through Compositionality}
\titlerunning{Enhancing Cross-modal Segmentation through Compositionality}
%
%
%
\author{Aniek Eijpe\inst{1,2,3}\orcidlink{0009-0009-7785-8885} \and
Valentina Corbetta \inst{2,3,4}\orcidlink{0000-0002-3445-3011} \and
Kalina Chupetlovska\inst{2}\orcidlink{0000-0002-1183-7426}\and
Regina Beets-Tan\inst{2,4}\orcidlink{0000-0002-8533-5090}\and
Wilson Silva\inst{2,3}\orcidlink{0000-0002-4080-9328}}

\authorrunning{A. Eijpe et al.}
%
\institute{University of Amsterdam, Amsterdam, The Netherlands\\
\email{\{aniek.eijpe\}@student.uva.nl}\and
Department of Radiology, The Netherlands Cancer Institute, Amsterdam, The Netherlands\and
AI Technology for Life, Department of Information and Computing
Sciences, Department of Biology, Utrecht University, Utrecht, The Netherlands\and 
GROW School for Oncology and Developmental Biology, Maastricht
University Medical Center, Maastricht, The Netherlands}

\maketitle              
\begin{abstract}
Cross-modal medical image segmentation presents a significant challenge, as different imaging modalities produce images with varying resolutions, contrasts, and appearances of anatomical structures.
We introduce compositionality as an inductive bias in a cross-modal segmentation network to improve segmentation performance and interpretability while reducing complexity.
The proposed network is an end-to-end cross-modal segmentation framework that enforces compositionality on the learned representations using learnable von Mises-Fisher kernels. These kernels facilitate content-style disentanglement in the learned representations, resulting in compositional content representations that are inherently interpretable and effectively disentangle different anatomical structures. The experimental results demonstrate enhanced segmentation performance and reduced computational costs on multiple medical datasets. Additionally, we demonstrate the interpretability of the learned compositional features. Code and checkpoints will be publicly available at: \url{https://github.com/Trustworthy-AI-UU-NKI/Cross-Modal-Segmentation}.

\end{abstract}
\section{Introduction}
Cross-modal medical image segmentation involves leveraging annotated images from one domain (e.g., \gls{ct}) to segment images in a different domain (e.g., \gls{mri}) without available segmentation labels. This domain shift~\cite{quinonero2008dataset} poses numerous challenges, as different imaging modalities produce images with varying resolutions, contrasts, and noise levels. Additionally, anatomical structures may appear differently or may not be visible across modalities due to varying imaging principles and acquisition parameters~\cite{ouyang2021representation}. This heterogeneity makes it difficult to develop a segmentation model trained on for example \gls{ct} that can effectively handle \gls{mri} as well.

In recent years, \gls{drl} techniques have found significant application in the context of cross-modal image segmentation, often employed in a cross-modal translation setup~\cite{wang2022cycmis,wang2021unsupervised,chen2021diverse,pei2021disentangle,chen2021beyond,xie2022unsupervised}. 
\gls{drl} aims to learn general and meaningful representations that capture the underlying factors that generate the data variation~\cite{bengio2013representation,higgins2018towards}. By disentangling (i.e. separating) these factors, the representations become inherently more interpretable and generalizable to other domains~\cite{wang2022disentangled}.  In the context of cross-modal segmentation, most works focus on \gls{csd}~\cite{gatys2016image}, separating the \textit{content} from the \textit{style} in two distinct representations. Generally, the style representations are considered domain-specific, while the content representations are viewed as domain-invariant and said to be shared across domains~\cite{liu2017unsupervised,lee2020drit++}, although these definitions are not universally agreed upon~\cite{liu2022learning}.  

Current methods require complex and large architectures and objective func- tions to bridge the gap between modalities [23,26,17,22], resulting in high com- putational costs. Moreover, defining and achieving proper and effective disen- tanglement between these representations is still a challenge. Multiple studies show that it is often unclear whether the learned representations are truly dis- entangled or not, showing information leakage between representations [16,13]. These challenges contribute to a lack of interpretability in current approaches.

Current methods require complex and large architectures and objective functions to bridge the gap between modalities~\cite{wang2022cycmis,xie2022unsupervised,pei2021disentangle,wang2021unsupervised}, resulting in high computational costs.
Moreover, defining and achieving proper and effective disentanglement between these representations is still a challenge. Multiple studies show that it is often unclear whether the learned representations are truly disentangled or not, showing information leakage between representations \cite{ouyang2021representation,liu2022learning}. These challenges contribute to a lack of interpretability in current approaches. 

We address these challenges by introducing compositionality as an inductive bias into an end-to-end cross-modal segmentation framework, aiming to reduce complexity and enhance interpretability. Compositionality refers to the notion that a representation as a whole should be composed of the representations of its parts~\cite{stone2017teaching}. This approach has been applied to numerous computer vision methods, demonstrating improvements in both robustness and explainability~\cite{tokmakov2019learning,kortylewski2020compositional,liu2022vmfnet}.
Our method enforces compositionality on the learned representations, which are modeled by learnable \gls{vmf} kernels~\cite{liu2022vmfnet}. These kernels effectively filter out style information. The resulting compositional, content representations contain only spatial information, additionally disentangling the anatomical structures from the background and each other, enhancing the interpretability of the learned representations. 
We argue that these content representations should not necessarily be shared across two different imaging modalities, since different modalities capture varying characteristics of anatomies~\cite{ouyang2021representation}. Relying solely on domain-invariant information for segmentation may lead to omitting crucial data, as some relevant information may be domain-specific. Therefore, we define the content representations as compositional rather than strictly domain-invariant.
By implicitly modeling the style representations within these kernels, we remove the need for domain-specific style encoders, thereby reducing the complexity and consequentially computational costs. 

The experimental results demonstrate enhanced performance and interpreta-\\bility with reduced computational costs on an unpaired public cardiac \gls{ct} \& \gls{mri} dataset and abdominal multi-modal \gls{mri} dataset.

Summarizing, our key contributions are the following:
\begin{itemize}
    \item[--] We introduce compositionality as an inductive bias into a cross-modal segmentation framework, effectively disentangling the content representations.
    \item[--] We reduce computational costs by implicitly modeling the style representations corresponding to several compositional components of human anatomy with learnable \gls{vmf} kernels.
    \item[--] We show enhanced cross-modal segmentation performance on two public, unpaired medical datasets and demonstrate the interpretability of the learned compositional content representations.
\end{itemize}

\section{Methodology}
\begin{figure}[th!]
     \centering
\includegraphics[width=0.5\textwidth]{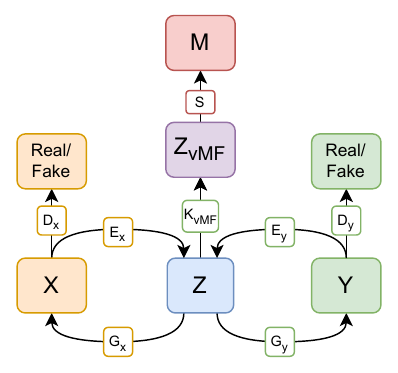}
     \caption{Overview of the proposed framework. $X$ and $Y$ denote the source and target domain from which the encoders $E_x$ and $E_y$ extract the deep features into $Z$. From $Z$, the deep features can be translated to either domain with the generators $G_x $ and $G_y$, or compositional representations $Z_{vMF}$ can be obtained via the vMF kernels ($K_{vMF}$). From $Z_{vMF}$, the segmentation model $S$ predicts the final segmentation masks. $D_x$ and $D_y$ denote the domain discriminators.}
    \label{fig:proposed}
\end{figure}

We aim to develop a model to segment images from a target domain $y \in Y$, by using images from a source domain $x \in X$ with corresponding labels $m_x \in M$. Figure \ref{fig:proposed} displays our proposed framework. 
In the context of \gls{csd}, we adopt a two-stage disentanglement approach.
Initially, our network roughly aligns the deep features $\mathbf{Z_x}$ and $\mathbf{Z_y}$ obtained by the two domain-specific encoders. Subsequently, the \gls{vmf} kernels $K_{vMF}$ filter out all remaining target-specific domain information to obtain the compositional, content representations $\mathbf{Z_{vMF}}$, which are used to obtain the final segmentation mask $\hat{m}_y \in M$.
\subsubsection{Cross-modal translation}
To obtain a mapping between the source and the target domain, the network is trained to perform cross-modal translation from the source to the target domain and vice versa.
$E_x$ and $E_y$ denote the domain-specific encoders, encoding the images of the source and target domain into deep features $\mathbf{Z_x}, \mathbf{Z_y} \in \mathbb{R}^{C_z \times H_z \times W_z}$, respectively. 
To translate the obtained representations to the other domain, the deep features $\mathbf{Z_x}$ and $\mathbf{Z_y}$, are used by the generators $G_y$ and $G_x$ respectively, to generate \textit{fake} images with the appearance of the respective domain but the content of the original domain. Since no paired data is available, a cross-cycle consistency loss~\cite{zhu2017unpaired} for both domains is used to learn the bi-directional mapping between the two domains.
\begin{equation}
    \mathcal{L}^{x,y}_{cycle} =  \mathbb{E}_{x,y} \left[|| G_{x,y}(E_{y,x}(G_{y,x}(E_{x, y}(\{x,y\})))) - \{x, y\} ||_1 \right]
\end{equation}

\noindent To enhance the quality of the generated images, an adversarial learning approach is adopted, with two domain discriminators $D_x$ and $D_y$, acting on the source and target domain, respectively. These discriminators are trained to distinguish between real and fake images, while the generators attempt to produce images that are indistinguishable from the real ones. The generator ($\mathcal{L}^{x,y}_{gen}$) and discriminator ($\mathcal{L}^{x,y}_{disc}$) losses are inspired by the Least Squares GAN (LSGAN)~\cite{mao2017least}. 
\begin{equation}
    \mathcal{L}^{x,y}_{gen} = \frac{1}{2} \mathbb{E}_{y,x} \left[ D_{x,y}(G_{x,y}(E_{y,x}(\{y, x\})))-1)^2 \right]
\end{equation}
\begin{equation}
    \mathcal{L}^{x,y}_{disc} = \frac{1}{2}\mathbb{E}_{x,y} \left[ (D_{x,y}(\{x,y\})-1)^2 \right] + \frac{1}{2} \mathbb{E}_{y,x} \left[ D_{x,y}(G_{x,y}(E_{y,x}(\{y,x\})))^2 \right]
\end{equation}

\subsubsection{Learning compositional representations}
\label{sec:learning}
\begin{figure}[h!]
     \centering
     \includegraphics[width=\textwidth]{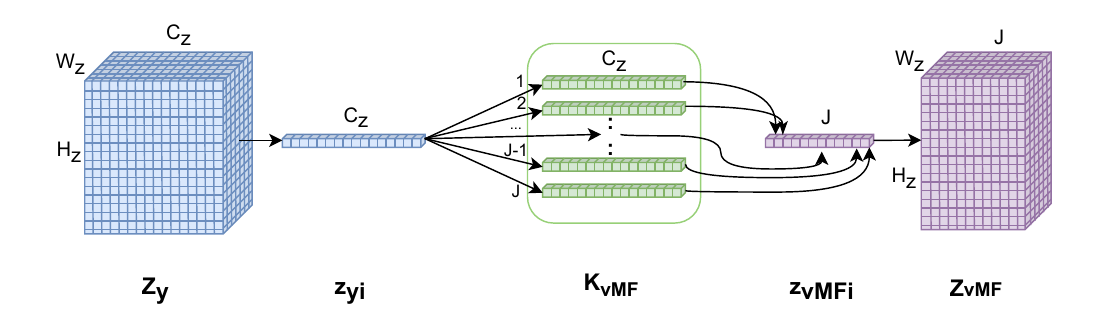}
     \caption{Visual overview of learning a compositional representation $\mathbf{Z_{vMF}}$ from the representation $\mathbf{Z_y}$ containing the deep features of a single target image $y$.}
    \label{fig:kerns_overview}
\end{figure}

With the deep features $\mathbf{Z_y}$ and the corresponding learnable \gls{vmf} kernels, $K_{vMF}$, we obtain the compositional representations $\mathbf{Z_{vMF}} \in \mathbb{R}^{J \times H_z \times W_z}$. This process is depicted in Figure \ref{fig:kerns_overview}.

$\mathbf{Z_y}$ can be viewed as a 2D map $H_z \times W_z$, with a feature vector $\mathbf{z_{yi}} \in \mathbb{R}^{C_z}$ on each position $i$. Following vMFNet's principles~\cite{liu2022vmfnet}, we model the feature representations $\mathbf{Z_y}$ as a mixture of $J$ von-Mises-Fisher distributions, where each distribution has a learnable mean. This learnable mean of distribution $j$ is referred to as the \gls{vmf} kernel $\boldsymbol{\mu}_{j} \in \mathbb{R}^{C_z}$. By calculating the vMF likelihood of each feature vector $\mathbf{z_{yi}}$ following each \gls{vmf} distribution $j \in \{1,..,J\}$, we obtain the compositional content representations $\mathbf{Z_{vMF}} \in \mathbb{R}^{J\times H_z \times W_z}$. The likelihood for each feature vector at position $i$ following the distribution $j$ is calculated as follows:

 \begin{equation}
 \label{eq1}
     p(\textbf{z}_{yi}|\boldsymbol{\mu}_{j}) =  C(\sigma)^{-1} \cdot e^{\sigma \boldsymbol{\mu}_{j}^T \textbf{z}_{yi}} 
 \end{equation}
\\
 \noindent where $||\boldsymbol{\mu}_{j}|| = 1$ and $||\textbf{z}_{yi}|| = 1$. For tractability, the variance $\sigma$ is fixed for all distributions. The obtained likelihoods are combined in $\mathbf{Z_{vMF}} \in \mathbb{R}^{J \times H_z \times W_z}$, indicating how much each kernel $j \in \{1,.., J\}$ is activated at each position $i$ in the 2D feature map $H \times W$. 
 
 During training, the network optimizes these kernels to serve as cluster centers of the feature vectors, using the following cluster loss:

\begin{equation}
    \label{eq:clu}
     \mathcal{L}^y_{vMF} =  \mathbb{E}_y \left[-(H \cdot W)^{-1}\sum_i \text{max}_j\boldsymbol{\mu}_{j}^T \textbf{z}_{yi} \right]
 \end{equation}

\noindent After training, the image feature vectors that are similar, activate the same kernels, have similar likelihoods, and are therefore clustered in the same channels of the compositional representation. These similar feature vectors are often induced by image patches that share semantic meaning~\cite{kortylewski2020compositional}.

\subsubsection{Segmentation}
\label{sec:seg}
The segmentation model $S$ is trained on the compositional content representations $\mathbf{Z_{vMF}}$, as these representations contain the spatial information. During training, only the labels of the source images are accessible. Therefore, we translate the source images to the target domain, and then from these translated images, the segmentation masks are obtained using the compositional features. The predicted segmentation masks are compared to the original segmentation labels of the source images using the Dice loss.
\begin{equation}
    \mathcal{L}_{seg}= \mathbb{E}_x \left[\textrm{Dice}(m_x, S(K_{vMF}(E_y(G_y(E_x(x))))))\right]
\end{equation}

\section{Experiments}
\subsubsection{Datasets}
From the \textbf{Multi-Modality Whole Heart Segmentation (MM-WHS)} challenge~\cite{zhuang2019evaluation} dataset, 320 \gls{ct} and 320 \gls{mri} (b-SSFP) slices were extracted\footnote{\url{https://github.com/FupingWu90/CT\_MR\_2D\_Dataset\_DA}} around the center of the left ventricular blood cavity along the long-axis view~\cite{wu2020cf} with annotated labels of the \gls{myo}, \gls{lv}, and \gls{rv}. 
From the \textbf{Combined Healthy Abdominal Organ Segmentation (CHAOS)} challenge~\cite{CHAOSdata2019,CHAOS2021}, multi-modal abdominal MR sequence data (T1-DUAL and T2-SPIR) of healthy patients with segmentation labels of the liver parenchyma were utilized, with 647 T1 and 623 T2 scans. This dataset contains substantial variation in image content, including negative examples (i.e., images without the liver). 

\subsubsection{Baselines} To evaluate the domain shift impact on segmentation performance, we employ a standard UNet~\cite{ronneberger2015u} in two ways. The \textbf{UNet-NA}, trained on source data and tested on target data, serves as a lower bound. The \textbf{UNet-FS} is fully supervised on target data, serving as an upper performance bound. The \textbf{vMFNet} model, robust to images collected from different clinical centres~\cite{liu2022vmfnet}, is tested against our large domain shift. The \textbf{DRIT+UNet} and \textbf{DRIT+RUNet} baselines use DRL in a two-stage cross-modal segmentation process: DRIT++~\cite{lee2020drit++} learns to translate images between domains, then UNet and Residual UNet~\cite{kerfoot2019left} are trained on synthetic target data. The \textbf{DDFSeg}~\cite{pei2021disentangle} model, modified to validate on source labels, is an end-to-end CSD network using domain invariant features for segmentation. This modification aligns with the principle of not using target ground truth labels during training.



\subsubsection{Implementation details}
The models are trained for 200 epochs using the Adam Optimizer \cite{kingma2014adam} with a batch size of 4, a learning rate of 0.0001, and exponential decay rates $(\beta_1, \beta_2) = (0.5, 0.999)$. 
Algorithm 1 in the supplementary materials outlines the exact training strategy of our model.
 We empirically chose to set the number of \gls{vmf} kernels to 10, which were initialized using Xavier initialization along with the rest of the network. Following Kortylewski \textit{et al.}~\cite{kortylewski2020compositional}, we fixed the variance of the \gls{vmf} distributions to 30. The model was trained and tested with 5-fold cross-validation, during which the model is validated with the \gls{dsc} and the \gls{lpips} metric~\cite{zhang2018unreasonable}, measuring both segmentation performance and quality of generated images.
 After training, the models are evaluated with the \gls{dsc} and the \gls{assd}.
\begin{table}[th!]
    \centering
    \caption{Quantitative comparison on the MM-WHS dataset showing the mean and std over the different folds. The best performance is denoted in \textbf{Bold}.}
    \begin{tabular}{|l|l l|l l|l l|}
        \hline
          \gls{mri} $\rightarrow$ \gls{ct}  & \gls{myo} & & \gls{lv} & & \gls{rv} & \\
          & \gls{dsc}(\%) & \gls{assd}(mm) & \gls{dsc}(\%) & \gls{assd}(mm) & \gls{dsc}(\%) & \gls{assd}(mm) \\
        \hline
         UNet-FS & $87.1_{2.4}$ & $1.3_{0.3}$ & $92.1_{1.1}$ & $1.4_{0.2}$ & $90.2_{2.4}$ & $1.8_{0.4}$ \\
        
        UNet-NA & $5.3_{4.7}$ & $26.7_{6.2}$ & $37.2_{23.4}$ & $19.7_{16.6}$ & $25.4_{22.5}$ & $22.9_{9.8}$ \\
        
        vMFNet & $2.3_{1.4}$ & $26.7_{4.0}$  & $52.4_{15.3}$  & $10.8_{3.7}$  & $40.4_{9.7}$ & $12.5_{1.3} $ \\
       
        DDFSeg & $24.5_{5.8}$ & $15.7_{4.7}$ & $39.1_{3.1}$ & $23.2_{1.3}$ & $24.2_{14.2}$ & $29.9_{2.7}$\\
    
        DRIT+UNet & $47.5_{8.5}$ & $5.3_{2.0}$ &$ 69.5_{3.3}$ & $6.0_{1.4}$ & $67.9_{5.8} $& $\mathbf{5.5_{0.8}}$ \\
        
        DRIT+RUNet & $58.4_{3.8}$ & $3.9_{0.2}$ & $75.1_{3.1}$ & $5.1_{0.5}$ & $71.5_{2.5}$ & $6.7_{1.5}$ \\
        
        Proposed  & $\mathbf{65.1_{4.8}}$ & $\mathbf{3.0_{0.6}}$ & $\mathbf{80.2_{4.7}}$ & $\mathbf{4.7_{1.5}}$ & $\mathbf{77.3_{3.6}}$ & $5.6_{2.0}$ \\
        \hline
    \end{tabular}
    \label{tab:res_mmwhs}
\end{table}

\subsubsection{Cardiac segmentation} Table \ref{tab:res_mmwhs} presents the quantitative results for segmenting the \gls{myo}, \gls{lv}, and \gls{rv} with CT as the target domain. The FS and NA baselines show a significant performance drop due to the domain shift. The vMFNet and DDF Segmentation network also perform poorly with this substantial modality shift. Our modification to the DDFSeg framework may have contributed to its reduced performance, as validating on true source labels provides a less precise estimate of the model's performance on the target domain compared to validating on true target labels.
Among the baselines, DRIT++ performs best, both with UNet and Residual UNet for second-stage segmentation. Overall, our proposed method outperforms all baselines. Supplementary Table 1 shows reverse direction results, and Figure 1 provides qualitative examples, demonstrating similar patterns. Introducing compositionality reduced training times significantly: 38 hours less than DRIT+UNet and DRIT+RUNet, and 16.5 hours less than DDFSeg on the MM-WHS dataset.

\subsubsection{Liver parenchyma segmentation} Table \ref{tab:chaos} presents the quantitative results for segmenting the liver parenchyma with T1 and T2 as target domain. Generally, we see a similar pattern as with the cardiac segmentation. Notably though, with T2 $\rightarrow$ T1, the drop in segmentation performance due to the domain gap is not as substantial. Our proposed method achieves the highest \gls{dsc} in both directions, but the \gls{assd} is substantially higher than the DRIT+UNet. Figure 1 in the supplementary material provides some qualitative results for both T1 target images and T2 target images.
\begin{table}[th!]
    \centering
    \caption{Quantitative comparison on the CHAOS dataset providing the mean and std over the different folds. The best performance is denoted in \textbf{Bold}.}
    \begin{tabular}{|l|l l|l l|}
    \hline
          Liver parenchyma   & T2 $\rightarrow$ T1 & & T1 $\rightarrow$ T2 &\\
          & \gls{dsc}(\%) & \gls{assd}(mm) & \gls{dsc}(\%) & \gls{assd}(mm) \\
        \hline
        UNet-FS & $89.1_{1.8}$ & $3.4_{0.9}$ & $88.5_{4.5}$ & $4.5_{1.9}$  \\
        
        UNet-NA & $62.9_{4.0}$ & $19.3_{7.7}$ & $25.2_{11.6}$ & $31.3_{8.5}$ \\

        vMFNet & $40.9_{9.5}$  & $29.5_{12.4}$ & $25.6_{9.6}$ & $22.0_{2.4}$ \\

        DDFSeg & $24.1_{11.7}$ & $69.3_{17.3}$ & $17.0_{9.3}$ & $69.6_{8.4}$ \\
       
        DRIT+UNet & $62.8_{5.9}$ & $\mathbf{17.8_{5.9}}$ & $54.1_{9.0}$ & $\mathbf{15.6_{3.3}}$  \\
        
        DRIT+RUNet & $58.8_{7.4}$ & $29.3_{9.3}$ & $53.9_{4.2}$ & $23.2_{2.4}$  \\ 
        
        Proposed  & $\mathbf{64.6_{4.6}}$ & $23.2_{5.3}$ & $\mathbf{66.3_{5.1}}$ & $19.1_{5.7}$ \\
        \hline
    \end{tabular}
    \label{tab:chaos}
\end{table}
\subsubsection{Compositional representations} 
\begin{figure}[t]
    \centering
     \begin{subfigure}{\textwidth}
         \includegraphics[width=\textwidth]{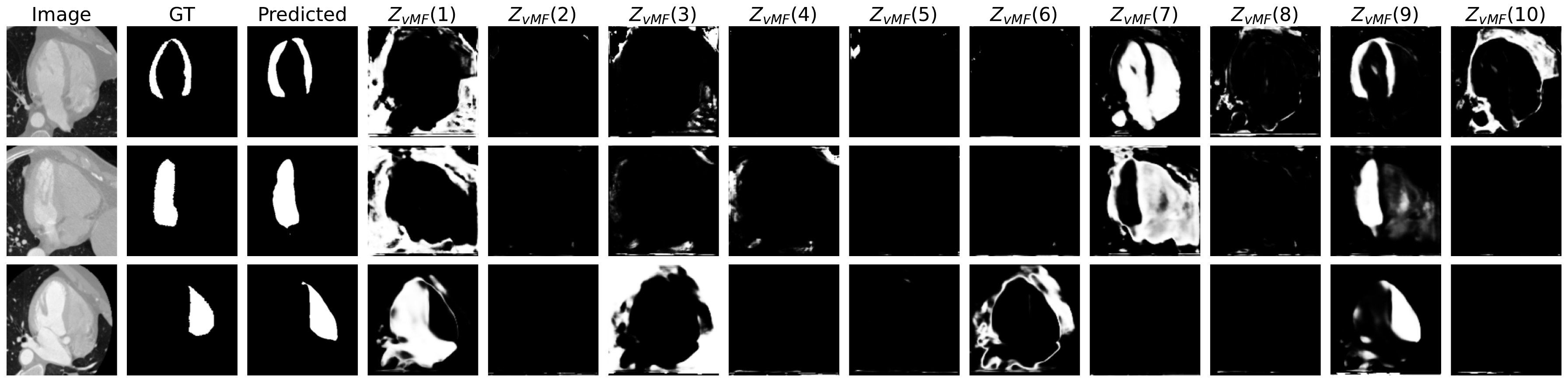}
     \end{subfigure} \\
     \begin{subfigure}{\textwidth}
         \includegraphics[width=\textwidth]{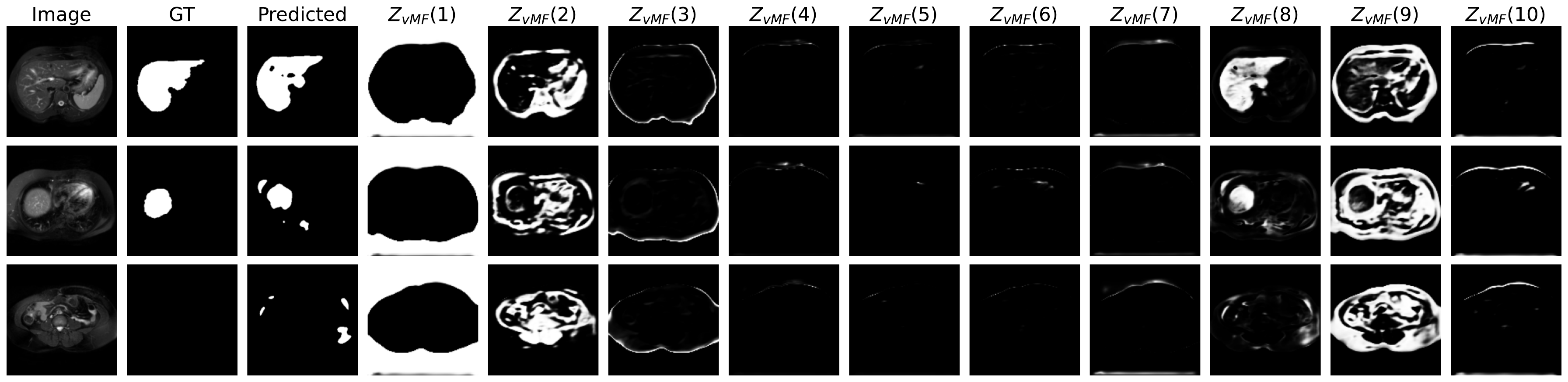}
     \end{subfigure}
    \caption{Visual results of our proposed method segmenting the \gls{myo}, \gls{lv}, \gls{rv}, with target \gls{ct} images, and the liver parenchyma with target T2-SPIR Images, with the 10 different channels of the compositional representation.}
    \label{fig:comp_repr}
\end{figure}

Figure \ref{fig:comp_repr} shows visual results of the learned compositional representations. The first three rows illustrate the network's ability to learn components representing anatomical structures like the \gls{myo}, \gls{lv}, \gls{rv}, other heart parts, and the chest wall. Despite missing some fine details, these interpretable representations achieve good segmentation performance. For liver parenchyma segmentation, a similar pattern is observed. However, the channels that are primarily activated by the liver parenchyma show merged activations. We suspect that due to the high variability in image content, the model struggles to find reliable style representations for the liver. Consequently, some non-liver patches activate channel 8, causing the network to produce a segmentation mask with several small artifacts even when no liver parenchyma is present in the image.
These artifacts elevate the \gls{assd} despite good \gls{dsc}. A post-processing step, retaining only the largest connected component in each predicted mask, can mitigate these artifacts.
\section{Conclusions}
We presented a novel end-to-end cross-modal segmentation framework that leverages images and segmentation labels from an annotated-rich domain to segment images from another, annotated poor domain. 
We introduced compositionality into a \gls{drl} network to address the lack of interpretability and high computational costs in the current models. By enforcing the learned representations to be compositional, we effectively disentangle style features from content features. These content features are further disentangled, separating the representations of different anatomical structures.
The qualitative and quantitative experiments demonstrated enhanced performance on cardiac \gls{ct}-\gls{mri} \gls{myo}, \gls{lv}, and \gls{rv} segmentation, outperforming cross-modal segmentation baselines employing disentangled representations. Moreover, our network showed an increase in the \gls{dsc} for cross-modal \gls{mri} liver parenchyma segmentation. Additionally, we reduced complexity, and thereby vastly decreased computational costs. Lastly, the interpretable nature of the learned compositional representations provided valuable insights into the segmentation process. Future research will focus on handling negative examples and expanding to multi-class segmentation to further improve compositional representations and consequently, segmentation performance.

\begin{credits}
\subsubsection{\ackname} Research at the Netherlands Cancer Institute is supported by grants from the Dutch Cancer Society and the Dutch Ministry of Health, Welfare and Sport. The authors would like to acknowledge the Research High Performance Computing (RHPC) facility of the Netherlands Cancer Institute (NKI). 

\subsubsection{\discintname}
The authors have no competing interests to declare that are relevant to the content of this article.
\end{credits}

%
%
\bibliographystyle{splncs04}
\bibliography{Paper-0025}

\end{document}